\documentclass{llncs}
\usepackage{subcaption}
\usepackage{graphicx}
\usepackage{xspace}
\usepackage{amssymb,amsmath}
\usepackage{ifthen}
\usepackage{multirow}
\usepackage{booktabs}
\usepackage{url}
\usepackage{listings}
\usepackage{totcount}
\usepackage{color, colortbl}

\newcommand{\act}[1]{\mathsf{#1}}

\newcommand{\wildch}{{\text{\_}}}

\newcommand{\sciff}{SCIFF\xspace}
\newcommand{\spec}{\ensuremath{\mathcal{S}}\xspace}
\newcommand{\abdset}{\ensuremath{\mathcal{A}}\xspace}
\newcommand{\kb}{\ensuremath{\Gamma}\xspace}
\newcommand{\ic}{\ensuremath{\mathcal{IC}}\xspace}
\newcommand{\trace}{\ensuremath{\mathcal{T}}\xspace}

\newcommand{\abd}[1]{\textsc{\textbf{#1}}}
\newcommand{\happened}{\abd{H}}
\newcommand{\ought}{\abd{E}}

\newcommand{\hyp}{\abd{ABD}}
\usepackage[defblank]{paralist}

\newcommand{\then}{\rightarrow}

\newcommand{\hapf}{\tau}
\newcommand{\expf}{\varepsilon}

\title{Abducing Compliance of Incomplete Event Logs}
\author{Federico Chesani\inst{1} \and Riccardo De Masellis\inst{2} \and Chiara Di Francescomarino\inst{2} \and Chiara Ghidini\inst{2}\and Paola Mello\inst{1} \and Marco Montali\inst{3} \and Sergio Tessaris\inst{3}}
\institute{University of Bologna, Bologna, Italy 
\and FBK-IRST, Via Sommarive 18, 38050 Trento, Italy
\and Free University of Bozen--Bolzano, piazza Universit\`{a}, 1, 39100 Bozen-Bolzano, Italy
{\fontsize{8}{10}\selectfont\email{\{federico.chesani,paola.mello\}@unibo.it}, \email{\{r.demasellis,dfmchiara,ghidini\}@fbk.eu}, \email{\{tessaris,montali\}@inf.unibz.it}} }

\begin{document}

\maketitle

\begin{abstract}
The capability to store data about business processes execution in so-called Event Logs has
brought to the diffusion of tools for the analysis of process executions
and for the assessment of the \emph{goodness} of a process model.
Nonetheless, these tools are often very rigid in dealing with with Event Logs
that include incomplete information about the process execution. Thus, while the ability of handling incomplete event data is one of the challenges mentioned in the process mining manifesto, the evaluation of compliance of an execution trace still requires an end-to-end complete trace to be performed.  
This paper exploits the power of abduction to provide a flexible, yet computationally effective, framework to deal with different forms of incompleteness in an Event Log. Moreover it proposes a refinement of the classical notion of compliance into \emph{strong} and \emph{conditional} compliance to take into account incomplete logs.  
Finally, performances evaluation in an experimental setting shows the feasibility
of the presented approach. 
\end{abstract}

\section{Introduction} 
\label{sec:introduction}

The proliferation of IT systems able to store process executions traces in so-called event logs has originated, in the Business Process community, a quest towards tools that offer the possibility of discovering, checking the conformance and enhancing process models based on actual behaviors~\cite{Aalstetal12}. Focusing on conformance, that is, on a scenario where the aim is to assess how a \emph{prescriptive} (or ``de jure'') process model relates to the execution traces, a fundamental notion os the one of \emph{trace compliance}. Compliance results can be used by business analysts to assess the goodness of a process model and understand how it relates to the actual behaviours exhibited by a company, consequently providing the basis for process re-design, governance and improvement.   

The use of event logs to evaluate the goodness of a process model becomes hard and potentially misleading when the event log contains only partial information on the process execution. Thus, while the presence of non-monitorable activities (or errors in the logging procedure) makes the ability of handling incomplete event data one of the main challenges of the BP community, as mentioned in the process mining manifesto\cite{Aalstetal12}, still trace compliance of an execution trace requires the presence of a complete end-to-end execution trace to be evaluated. Notable exceptions are~\cite{Adriansyahetal:2011,deLeonietal:2012} where trace incompleteness is managed in an algorithmic/heuristic manner using log repair techniques.     

In this paper, we take an orthogonal approach and throughly address the problem of log incompleteness from a formal/logic-based point of view, adopting an approach based on \emph{abduction} \cite{DBLP:journals/logcom/KakasKT92}. Differently from techniques that focus on algorithmic/heuristic repairs of an incomplete trace, we are interested in characterising the notion of incomplete log compliance by means of a sound and complete inference procedure.  
We rely on abduction to combine the partial knowledge about the real executions of a process as reflected by a (potentially) incomplete event log, with the background knowledge captured in a process model. In particular, abductive reasoning handles different forms of missing information by formulating \emph{hypotheses} that explain how the event log may be ``completed'' with the missing information, so as to reconcile it with the process model. This leads us to refine the classical notion of \emph{conformance-by-alignment} \cite{Vda11-mining} between an execution trace and a process model into \textbf{strong} and \textbf{conditional} compliance, to account for incompleteness. 
In detail, the paper provides: (i) a classification of different forms of incompleteness of an event log based on three dimensions: log incompleteness, trace incompleteness, and event description incompleteness (Section~\ref{sec:classifying_log_incompleteness}); (ii) a reformulation of the notion of compliance into strong and conditional compliance (Section~\ref{sec:compliance}); (iii) an encoding of structured process models\footnote{We focus on structured process models in the spirit of~\cite{kiepuszewskihofstedeetal:2013}. Broadly speaking, this restricts to the class of models recursively composed of single-entry-single-exit blocks, where every split has a corresponding join, matching its type. This assumption rules out pathological patterns that are notoriously hard to characterise (e.g.\ involving nested OR joins), still providing coverage for a wide range of interesting use cases.} and of event logs in the SCIFF abductive logic framework~\cite{DBLP:journals/tocl/AlbertiCGLMT08}, and a usage of the SCIFF abductive proof procedure to compute strong, conditional and non- compliance in presence of an incomplete event log (Section~\ref{sec:approach}); and (iv) an evaluation of the proposed framework in an experimental setting (Section~\ref{sec:evaluation}). 
The ideas presented in the paper paper are illustrated by means of a simple explanatory example, and the comparison with related work is contained in Section~\ref{sec:related_works}.


\section{Dealing with Incomplete Event Logs} 
\label{sec:dealing_with_incomplete_event_logs}

We aim at solving solve the problem of identifying compliant traces in the presence of incomplete event logs, given the \emph{prescriptive} knowledge contained in a process model. 
To do this, we first need to investigate what incomplete event logs are (Section~\ref{sec:classifying_log_incompleteness}) and then understand how we can adapt the notion of compliance to deal with partial data on the process execution (Section~\ref{sec:compliance}).
We perform this investigation with the help of a simple example, which in this paper is described using the BPMN language\footnote{For the sake of clarity we use BPMN, but our framework is language-independent.}. 

\begin{figure}[t]
  \centering
    \includegraphics[width=.85\textwidth]{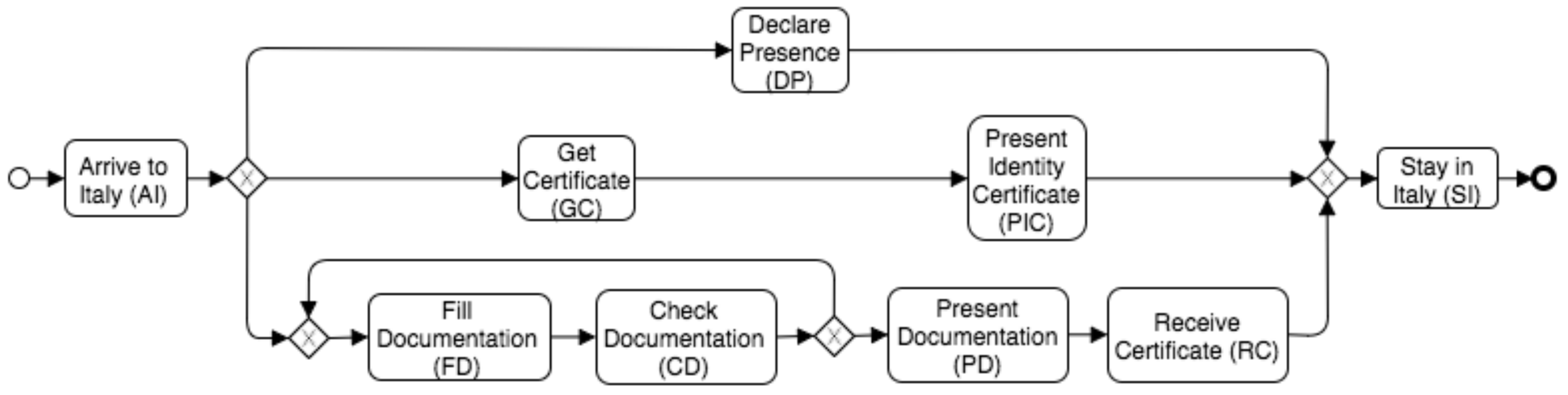}
  \caption{A process for obtaining a permit of stay (in Italy).}
  \vspace{-.5cm}
  \label{fig:run_ex}
\end{figure}

\begin{example}[Obtaining a Permit of Stay in Italy]
	\label{ex:POS}
	Consider the BPMN process in Figure~\ref{fig:run_ex}, hereafter called the Permit-Of-Stay (POS) process, which takes  inspiration from the procedure for the granting of a permit of stay in Italy.
	Upon her arrival in Italy ($\act{AI}$), the person in need of a permit of stay has three different alternatives: 
	if she is from a EU country and remains in Italy for at most 30 days, then only indicating her presence in Italy ($\act{DP}$) is needed;
	if she is from the EU and must remain in Italy for more than 30 days, then she needs to get an identity certificate ($\act{GIT}$) and present it ($\act{PIC}$). In all the remaining cases she needs to fill a documentation ($\act{FD}$) which is then checked ($\act{CD}$). When the documentation is correct, it is presented ($\act{PD}$) and a certificate is received ($\act{RC}$).
	The procedure concludes with the provision of the permit of stay ($\act{SI}$). Note that, for the sake of simplicity, the process only focuses on the so-called ``happy paths'', that is, the successful issuing of a permit of stay.  
\end{example}

\subsection{Classifying Process Execution (In)Completeness} 
\label{sec:classifying_log_incompleteness}

We assume that each execution of the POS process in Figure~\ref{fig:run_ex} is (partially) monitored and logged by an information system.  
We also assume that activities are atomic, i.e., executing an activity results in an event associated to a single timestamp: event $(\act{A},t)$ indicates that activity $\act{A}$ has been executed at time $t$.
A sample trace\footnote{We often present the events in a trace ordered according to their execution time. This is only to enhance readability since the position of an event is fully determined by its timestamp, or unknown if the timestamp is missing.} that logs the execution of a POS instance is:
\begin{equation}
	\label{eq:sample-trace}
		\{(\act{AI}, t1), (\act{FD}, t2), (\act{CD}, t3), (\act{PD}, t4), (\act{RC}, t5), (\act{SI}, t6)\}
\end{equation}
where $t_i > t_j$ for $i,j\in\{1,\ldots,6\}$ such that $i > j$. This trace corresponds to the execution of the lower branch of the POC process, where the loop is never executed. A set of execution traces of the same process form an event log. 		

In many real cases, a number of difficulties may arise when exploiting the data contained in an information system in order to build an event log. For instance, data may bring only partial information in terms of which process activities have been executed and what data or artefacts they produced. Thus, instead of the extremely informative trace reported in \eqref{eq:sample-trace}, we may obtain something like:
\begin{equation}
	\label{eq:incomplete-trace}
		\{(\act{FD}, \_), (\_,t2), (\act{SI}, t6), (\_, \_)\}
\end{equation}
This trace does not completely describe an execution of the POS process. For example, the first event logged in the trace is $\act{FD}$. 
However, by looking at the process description, it is easy to see that the first event of every execution has to be $\act{AI}$. By assuming that the process executors indeed followed the prescriptions of the model, this suggests that the $\act{AI}$-related event has not been logged.  Moreover, certain events have been only partially observed. For example, the $\act{FD}$-related event is incomplete, because its exact timestamp is unkown. In this paper, we use ``\_'' to denote a missing information unit. 

\begin{figure}[t]
\begin{subfigure}[b]{.5\linewidth}
\centering\large 
    \includegraphics[width=\textwidth]{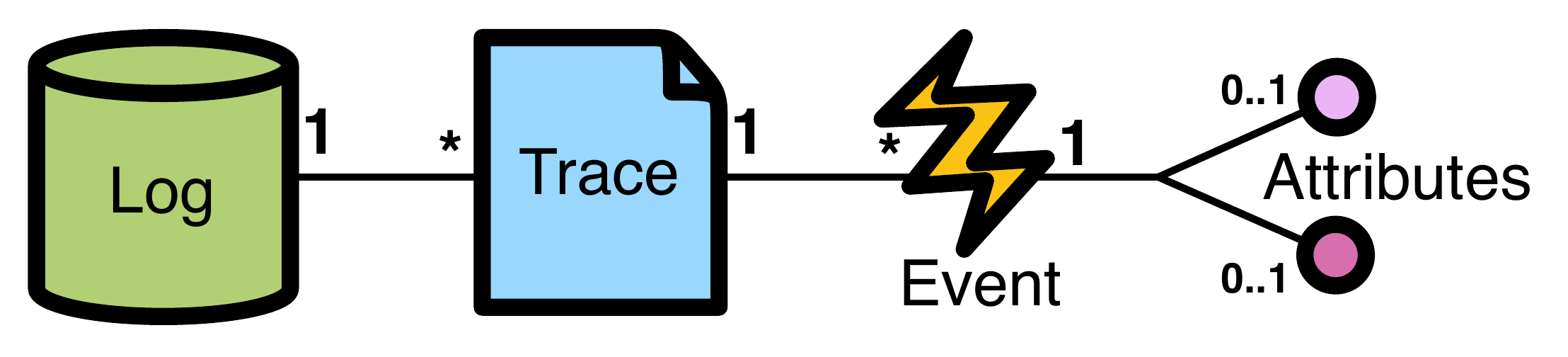}
  \caption{The Event Log Structure}
  \label{fig:figure_incompleteSchema}
\end{subfigure}\\[-1.6cm]
\begin{subfigure}[b]{.5\linewidth}
\centering\large \includegraphics[width=\textwidth]{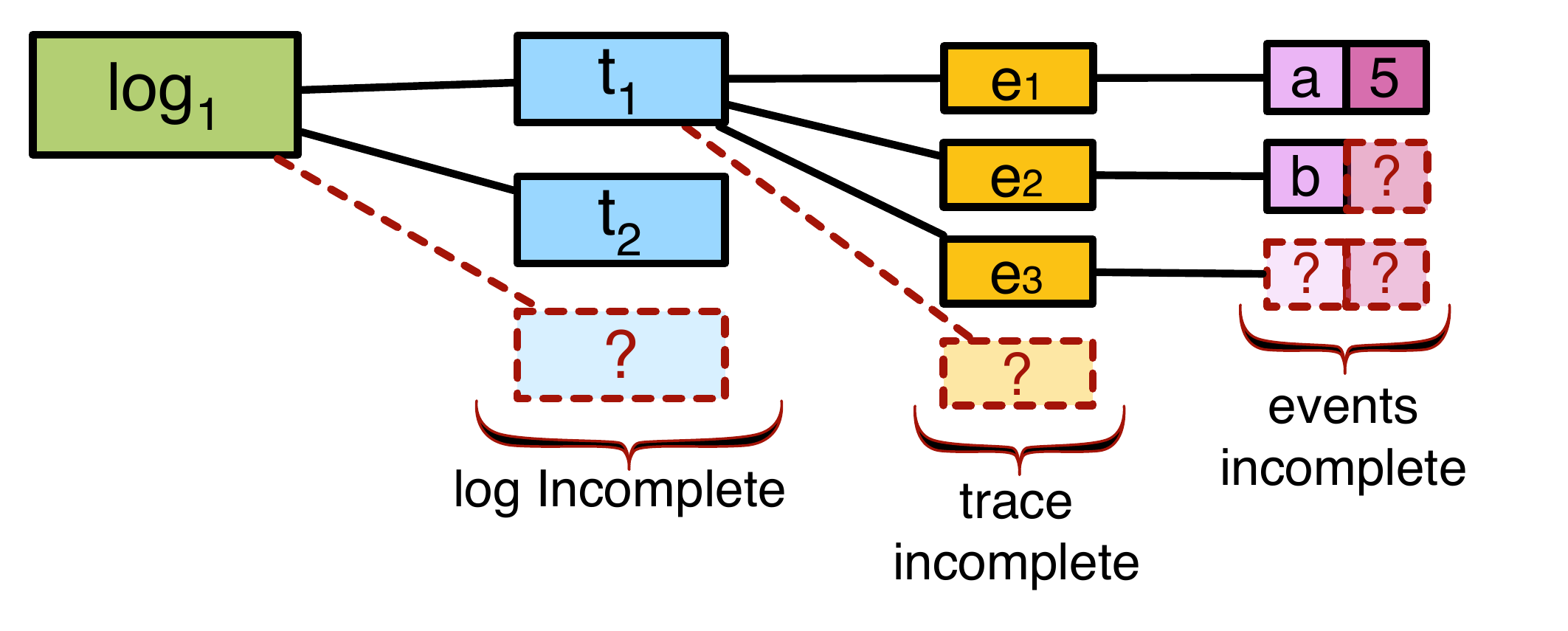}
  \caption{Three Incompleteness Dimensions}
  \label{fig:figure_incompleteSamples}
\end{subfigure}
\begin{subfigure}[b]{.4\linewidth}
\centering\large \includegraphics[width=\textwidth]{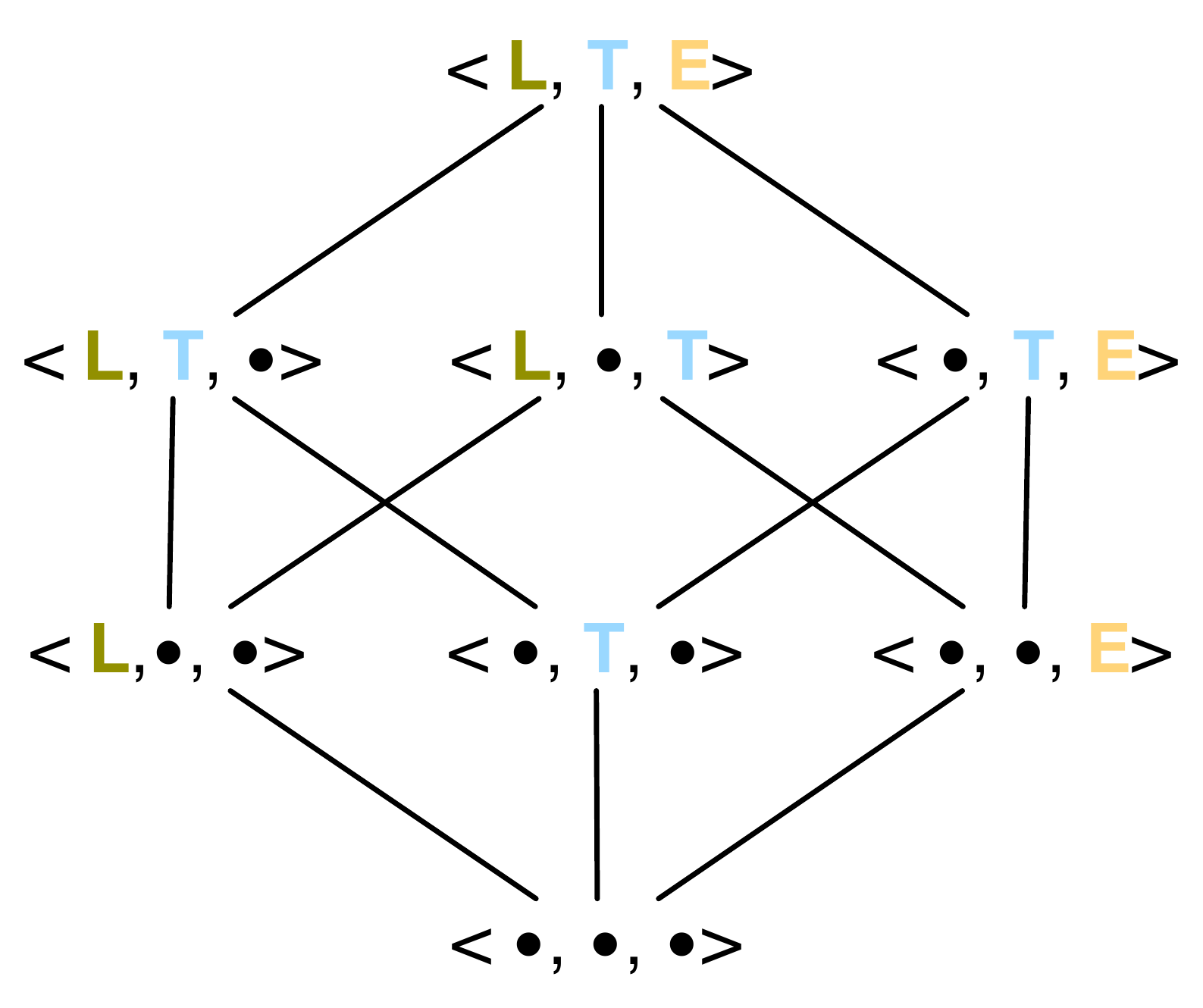}
  \caption{Combining Dimensions}
  \label{fig:figure_Incompleteness3}  
\end{subfigure}
\caption{Classifying (in)completeness.}\label{fig:incompleteness}
\vspace{-.5cm}
\end{figure}

In accordance with the IEEE standard XES format for representing event logs \cite{xes}, in general we can describe an event log as a set of execution traces. Each trace, in turn, contains events, which are described by means of n-tuples, where each element of the tuple is the value of a given attribute (see Figure \ref{fig:figure_incompleteSchema}, where we restrict to two attributes as we do in the paper). Consequently, we can classify incompleteness along these three dimensions: incompleteness of the log, incompleteness of the trace, and incompleteness of the event description (see Figure \ref{fig:figure_incompleteSamples}). 

\paragraph{(In)Completeness of the log.}
Within this dimension we analyse whether all the traces envisaged by the model are in the log or not. That is, we focus on understanding whether the log contains at least one instance for each possible execution that is allowed in the model. 
		Note that one can account for this form of (in)completeness only by: (a) limiting the analysis to the control flow, without considering complex data objects that may contain values from an unbounded domain; and (b) assuming that there is a maximum length for all traces, thus limiting the overall number of traces that may originate from the unbounded execution of loops.
An example of complete log for the POS process is: 	
	 \begin{equation}
		\label{eq:completeLog} 
		        L_1 = 
\left\{
\begin{array}{@{}l@{}}
  \{(\act{AI}, t_{a1}), (\act{DP}, t_{a2}), (\act{SI}, t_{a3})\},\\
				 \{(\act{AI}, t_{b1}), (\act{GIC}, t_{b2}), (\act{PIC},t_{b3}) (\act{SI}, t_{b4})\},\\
		        	 \{(\act{AI}, t_{c1}), (\act{FD}, t_{c2}), (\act{CD}, t_{c3}), (\act{PD}, t_{c4}), (\act{RC}, t_{c5}),
(\act{SI},t_{c6})\} \}
\end{array}
\right\}
	 \end{equation}
	 where we assume that each trace cannot contain more than 6 event, which intuitively means that the loop is never executed twice.
  
Assuming this form of strict completeness is often unrealistic in practice. In fact, even under the assumption of a maximum trace length, the number of allowed traces could become extremely huge due to (bounded) loops, and the (conditional) interleavings generated by parallel blocks and or choices. Still, analysing the (in)completeness of an event log may be useful to discover parts of the control flow that never occur in practice. 

\paragraph{(In)completeness of the trace.} Within this dimension we focus on a single trace, examining whether it contains a sequence of events that corresponds to an execution foreseen by the process model from start to end.  Trace \eqref{eq:sample-trace} is an example of complete trace. An example of incomplete trace is: 
 \begin{equation}
	\label{eq:incompleteTrace} 
		\{(\act{AI}, t_{1}),  (\act{PIC},t_{2}) (\act{SI}, t_{3})\}
 \end{equation}
By looking at the POS model, it is easy to see that this trace should also contain an event of the form $(GIC,t)$, s.t.~$t_1 < t < t_2$. 
%
%

\paragraph{(In)completeness of the event description.} Within this dimension we focus on the completeness of a single event. Events are usually described as complex objects containing data about the executed activity, its time stamp, and so on \cite{xes}. These data can be missing or corrupted. As pointed out before, we consider activity names and timestamps. Thus, incompleteness in the event description may concern the activity name, its timestamp, or both. This is reflected in trace \eqref{eq:incomplete-trace}: 
\begin{inparaenum}[\it (i)]
\item event $(\act{FD}, \_)$ indicates that activity $\act{FD}$ has been executed, but at an unknown time;
\item  $(\_,t2)$ witnesses that an activity has been executed at time $t_2$, but we do not know which;
\item  $(\_, \_)$ attests that the trace contains some event, whose activity and time are unknown.
\end{inparaenum}

\smallskip
\noindent
In general, we can characterise the (in)completeness of an event log in terms of (any) combination of these three basic forms. At one extreme, we may encounter a log that is complete along all three dimensions, such as the one depicted in \eqref{eq:completeLog}.
At the other extreme, we may instead have the following log:
\begin{equation}
	\begin{split}
		\label{eq:incompleteLog}
		L_2= \{ & \{(\act{AI},\wildch ),(\wildch,t_{a2})\},  \{(\act{AI}, t_{b1}),(\wildch,\wildch), (\wildch, t_{b2}), (\act{SI}, t_{b3}) \}
		\}
	\end{split}
\end{equation}
characterised by incompleteness of the log, incompleteness of some traces, and incompleteness of some event descriptions. Intermediate situations may obviously arise as well. This is graphically depicted in the lattice of Figure~\ref{fig:figure_Incompleteness3}, where $\langle L,T,E \rangle$ indicates the top value (completeness for all three dimensions) and $\langle \bullet,\bullet,\bullet \rangle$ indicated the bottom value (incompleteness of all three dimensions).  

\subsection{Refining the Notion of Compliance} 
\label{sec:compliance}

In our work we consider \emph{prescriptive} process models, that is, models that describe the only acceptable executions. These corresponds to the so-called ``de jure'' models in \cite{Vda11-mining}, and consequently call for a definition of \emph{compliance}, so as to characterise the degree to which a given trace conforms/is aligned to the model. The traditional notion of compliance is typically considered under the assumption that the trace is a faithful footprint of reality, and requires that the trace represents an end-to-end, valid execution that can be fully replayed on the process model. We call this notion of compliance \textbf{strong} compliance. Trace~\eqref{eq:sample-trace} is an example of trace that is fully compliant to the POS process.

Strong compliance is too restrictive when the trace is possibly incomplete. 
%
%
In fact, the incompleteness in a trace hinders the possibility of replaying it on the process model. However, full conformance might be regained by assuming that the trace included additional activities and/or specific information on the missing data; in this case we say that the trace is \textbf{conditionally} compliant, to reflect that compliance conditionally depends on how the partial information contained in the trace is complemented with the missing one.  Consider again the POS example and the partial trace:
 \begin{equation}
	\label{eq:abductedtrace}
		\{(\act{AI}, t_{1}),  (\act{GIC},\_) (\act{SI}, t_{3})\}
 \end{equation}
 It is easy to see that the observed trace is compliant with POS, \emph{if} we assume that
 \begin{gather}
	 \label{eq:h1}
	\act{GIC} \text{ was executed at a time } t_i \text{ s.t.~} t_1 < t_i < t_3\\
	\label{eq:h2}
	\text{an execution of } \act{PIC} \text{ was performed at a time } t_j \text{ s.t.~} t_i < t_j < t_3
\end{gather}
Note that the set of assumptions needed to reconstruct full conformance is not necessarily unique. This reflects that, in general, alternative strongly compliant real process executions might have led to the recorded partial trace. On the other hand, there are situations in which it is not possible to formulate additional assumptions on the partial trace to recover full conformance. In this case, the partial trace is considered \textbf{non}-compliant. 
For example, trace
\begin{equation}
	\label{eq:noncompliant}
	\{(\act{AI}, t_{1}),  (\act{GIC},\_) (\act{CD},t_{2}) (\act{SI}, t_{3})\}
\end{equation}
does not comply with POS, since it records that $\act{GIC}$ and $\act{CD}$ have been both executed, although they belong to mutually exclusive branches in the model.


\section{Abduction and Incomplete Logs} 
\label{sec:approach}
 
Since the aim of this paper is to provide automatic procedures, embedded in a tool, that identify compliant traces in the presence of incomplete event logs, given the \emph{prescriptive} knowledge contained in a process model, we can schematise the input to our problem in three parts: (i) an instance-independent component, the process model, which in this paper is described using BPMN; (ii) an instance-specific component, that is, the (partial) log, and (iii) meta-information attached to the activities in the process model, indicating which are actually \emph{always}, \emph{never} or \emph{possibly} observable (that is, logged) in the event log. The third component is an extension of a typical business process specification that we propose (following and extending the approach described in \cite{CAiSE2015}) to provide \emph{prescriptive} information about the (non-) observability of activities. Thus, for instance, a business analyst will have the possibility to specify that a certain manual activity is never observable while a certain interaction with a web site is always (or possibly) observable. This information can then be used to compute the compliance of a partial trace. In fact the presence of never observable activities will trigger the need to make hypothesis on their execution (as they will never be logged in the event log), while the presence of always observable activities will trigger the need to find their corresponding event in the execution trace (to retain compliance). Note that this extension is not invasive w.r.t. current approaches to business process modelling, as we can always assume that a model where no information on observability is provided is entirely possibly observable.


Given the input of our problem, we structure this section as follows: first, we provide an overview on abduction and on how the \sciff framework represents \emph{always}, \emph{never} or \emph{possibly} observable activities; then, we show how to use \sciff to encode a process model and a partial log (Section~\ref{subsec:yawlInSciff}), third we show how we can formalize, and therefore make precise, the informal different forms of compliance presented in Section~\ref{sec:compliance} (Section~\ref{subsec:sciffSemantics}); finally, we illustrate how the \sciff proof procedure can be used to solve the different forms of incompleteness identified in Section~\ref{sec:classifying_log_incompleteness} (Section~\ref{subsec:trace_incompleteness}).  

\subsection{The \sciff in short} 
\label{sub:the_sciff_in_short}
Abduction is a non-monotonic reasoning process where hypotheses are made to explain observed facts \cite{DBLP:conf/iclp/KakasM94}. While deductive reasoning focuses on deciding if a formula $\phi$ logically follows from a set $\Gamma$ of logical assertions known to hold, in abductive reasoning it is assumed that $\phi$ holds (as it corresponds to a set of observed facts) but it cannot be directly inferred by $\Gamma$. To make $\phi$ a consequence of $\Gamma$, abduction looks for a further set $\Delta$ of hypothesis, taken from a given set of abducible $\abdset$, which complements $\Gamma$ in such a way that $\phi$ can be inferred (in symbols $\Gamma \cup \Delta \models \phi$). The set $\Delta$ is called \emph{abductive explanation} (of $\phi$). 
In addition, $\Delta$ must usually satisfy a set of (domain-dependent) integrity constraints \ic (in symbols, $\Gamma \cup \Delta \models \ic$). A typical integrity constraint (IC) is a \emph{denial}, which expresses that two explanations are mutually exclusive. 

Abduction has been introduced in Logic Programming in \cite{DBLP:journals/logcom/KakasKT92}. There, an \emph{Abductive Logic Program (ALP)}  is defined as a triple $\langle \kb, \abdset, \ic \rangle$, where:
\begin{inparaenum}[\it (i)] 
\item $\kb$ is a logic program, 
\item $\abdset$ is a set of abducible
predicates, and 
\item $\ic$ a set of ICs.
\end{inparaenum}
Given a goal
$\phi$, abductive reasoning looks for a set of literals $\Delta\subseteq\abdset$ such that they entail $\phi \cup \ic$. 

In this paper we leverage on the SCIFF abductive logic programming framework~\cite{DBLP:journals/tocl/AlbertiCGLMT08}, an
extension of the IFF abductive proof procedure~\cite{DBLP:journals/jlp/FungK97}.
Beside the general notion of abducible, the \sciff\
framework has been enriched with the notions of \emph{happened
event}, \emph{expectation}, and \emph{compliance} of an observed execution with a set of
expectations. This makes \sciff\  suitable for dealing with event log incompleteness. 
Let $\act{a}$ be an event corresponding to the execution of a process activity, and $T$ (possibly with subscripts) its execution time\footnote{In the remainder of this paper we will assume that the time domain relies on natural numbers.}. 
Abducibles are used here to make hypothesis on events that are not recorded in the examined trace. They are denoted using $\hyp(\act{a},T)$. 
Happened events are non-abducible, and account for events that have been logged in the trace. They are denoted with $\happened(\act{a},T)$. 
Expectations $\ought(\act{a},T)$, instead, model events that should occur (and therefore should be present in a trace).
Compliance is described in Section~\ref{subsec:sciffSemantics}. 

ICs in \sciff\ are used to relate happened events / abduced predicates with expectations / predicates to be abduced. Specifically, an IC is a rule of the form $body \rightarrow head$, where $body$ contains a conjunction of happened events, general abducibles, and defined predicates, while $head$ contains a disjunction of conjunctions of expectations, general abducibles, and defined predicates.


\subsection{Encoding Structured Processes and Their Executions in \sciff}
\label{subsec:yawlInSciff}

Let us illustrate how to encode all the different components of an (incomplete) event log and a structured process model one by one. 

\vspace{-.3cm}
\paragraph{Event Log.}
A log is a set of traces, each constituted by a set of observed (atomic) events.
Thus trace~\eqref{eq:incompleteTrace} is
represented in \sciff\ as 
$\{\happened(\act{AI}, t_{1}),  \happened
(\act{PIC},t_{2}), \happened(\act{SI}, t_{3})\}.$

\begin{figure}[t]
	\centering
\begin{subfigure}[b]{.35\linewidth}
\centering
    \includegraphics[width=\textwidth]{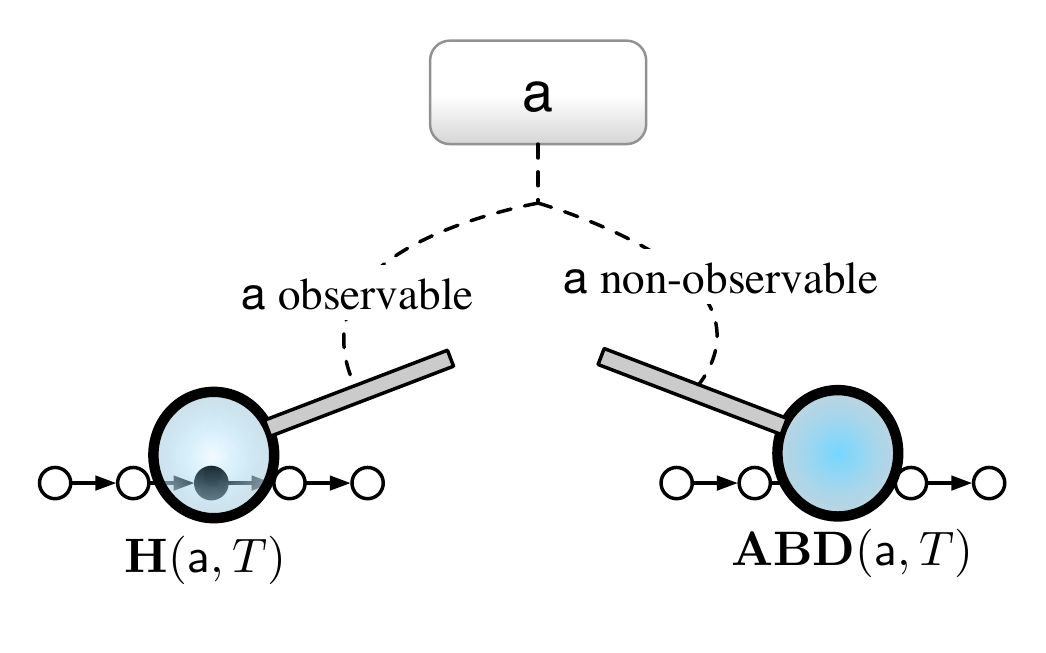}
    \vspace{-.35cm}
  \caption{Encoding execution with $\hapf$.}
  \label{fig:figure_alwaysObs}
\end{subfigure} \qquad
\begin{subfigure}[b]{.35\linewidth}
\centering \includegraphics[width=\textwidth]{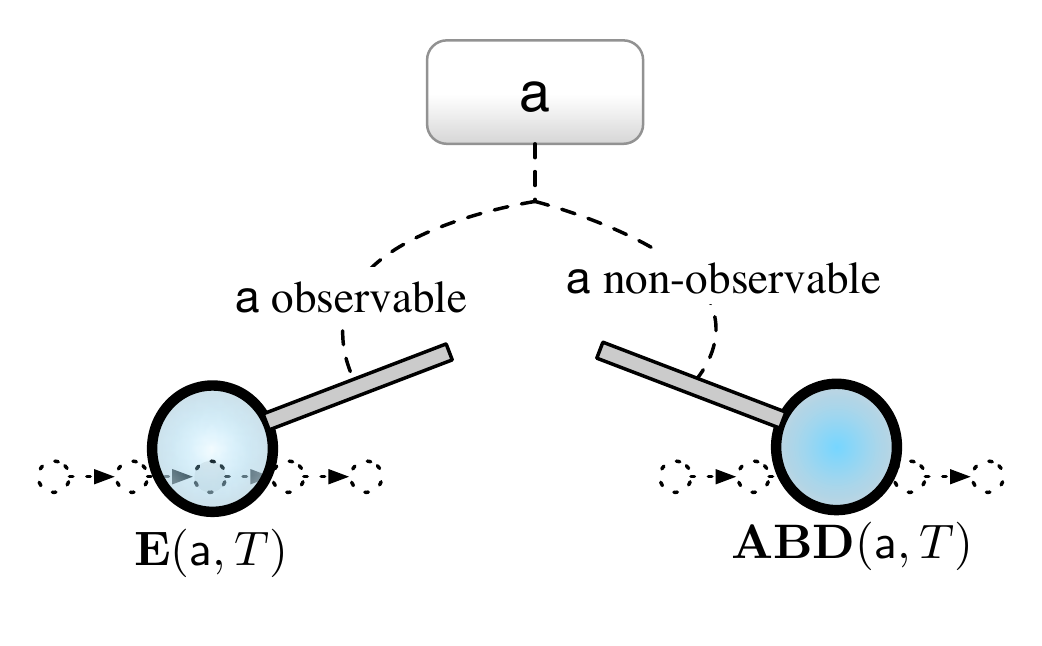}
\vspace{-.5cm}
  \caption{Encoding expectation with $\expf$.}
  \label{fig:figure_neverobs}
\end{subfigure}
\caption{Encoding always/never observable activities.}\label{fig:obs}
\vspace{-.5cm}
\end{figure}

\vspace{-.3cm}
\paragraph{Always/never observable activities.}
Coherently with the representation of an execution trace, the logging of the execution of an observable activity is 
represented in \sciff\ using an happened event, whereas the hypothesis on the execution of a never observable activity is represented using an abducible $\hyp$ (see Figure~\ref{fig:figure_alwaysObs}). 
Given an event $\act{a}$ occurring at $T$, we use a function $\hapf$
that represents the execution of $\act{a}$ as:
$$
\hapf(\act{a},T) = 
\begin{cases}
\happened(\act{a},T) & \text{if $\act{a}$ is observable}\\
\hyp(\act{a},T) & \text{if $\act{a}$ is never observable}
\end{cases}
$$ 
As for expected occurrences, the encoding again depends on the observability
of the activity: if the activity is observable, then its expected occurrence
is mapped to a \sciff\ expectation; otherwise, it
is hypothesised using the aforementioned abducible
$\hyp$ (see Figure~\ref{fig:figure_neverobs}). To this end we use a function $\expf$ that maps the expecting of the execution of $\act{a}$ at time $T$ as follows:
\[
\expf(\act{a},T) = 
\begin{cases}
\ought(\act{a},T) & \text{if $\act{a}$ is observable}\\
\hyp(\act{a},T) & \text{if $\act{a}$ is never observable}
\end{cases}
\]

\paragraph{Structured process model constructs.}

A process model is encoded in \sciff\ by generating ICs that relate
the execution of an activity to the future, expected executions of
further activities. In practice, each process model construct is transformed
into a corresponding IC. We handle, case-by-case, all the single-entry single-exit block types of
structured process models.

\smallskip
\noindent
\textbf{Sequence.}
Two activities $\act{a}$ and $\act{b}$ are in sequence if, whenever the first is
executed, the second is expected to be executed at a later time:
\begin{gather}
	\label{eq:encoding-sequence}
\hapf(\act{a},T_a) \then \expf(\act{b},T_b) \land T_b > T_a.
\end{gather}

\smallskip
\noindent
\textbf{Xor-split} extends sequence with the possibility of
selecting one among multiple target activities. In \sciff, this is
captured using an IC with a disjunction in the consequent. In
particular, the fact that if $\act{a}$ is executed, then either $\act{b}$ or $\act{c}$
is expected to be executed next is encoded as:
\begin{equation*}
\hapf(\act{a},T_a) \then 
\expf(\act{b},T_b) \land T_b > T_a 
\lor
\expf(\act{c},T_c) \land T_c > T_a.
\end{equation*}

\smallskip
\noindent
\textbf{Xor-join} indicates that, no matter which activity is executed
among the input set of activities for the xor-join, then the output
activity of the xor-join is expected to be executed. Hence, the
encoding of xor-join can be obtained by encoding all its implied
sequences. For example, if $\act{a}$ or $\act{b}$ are followed by
$\act{c}$, we obtain:
\begin{align*}
& \hapf(\act{a},T_a) \then \expf(\act{c},T_c) \land T_b > T_a.
& \hapf(\act{b},T_b) \then \expf(\act{c},T_c) \land T_c > T_b.
\end{align*}

\noindent
\textbf{And-split} activates parallel threads spanning from the same
activity. In particular, the fact that activity $\act{a}$ triggers two
parallel threads, one expecting the execution of $\act{b}$, and the
other that of $\act{c}$, is captured using an IC with a
conjunctive consequent:
\begin{equation*}
\hapf(\act{a},T_a) \then 
\expf(\act{b},T_b) \land T_b > T_a 
\land
\expf(\act{c},T_c) \land T_c > T_a.
\end{equation*}

\noindent
\textbf{And-join} mirrors the and-split, synchronizing multiple
concurrent execution threads and merging them into a single thread. When activities $\act{a}$ and $\act{b}$ are both
executed, then activity $\act{c}$ is expected next, is
captured using an IC with a conjunctive antecedent:
\begin{equation*}
\hapf(\act{a},T_a) 
\land
\hapf(\act{b},T_b)
\then 
\expf(\act{c},T_c) \land T_c > T_a \land T_c > T_b.
\end{equation*}

\noindent
\textbf{Or-split/or-join} are captured by combining the
formalization of and-/xor-elements, considering the well-known
equivalence between an or-split/-join and an xor-split/-join whose
alternative branches correspond to an element in the powerset of the
split-targets/join-sources, whose inner activities are put in
parallel. For example, the fact that $\act{a}$ leads to an or-split
pointing to $\act{b}$ and $\act{c}$ is equivalent to an xor-split that
connects $\act{a}$ to three outputs: one containing $\act{b}$, one
containing $\act{c}$, and one containing $\act{b}$ and $\act{c}$ in
parallel.

\paragraph{Possibly observable activities.}
A possibly observable activity is managed by considering
the disjunctive combination of two cases: one in which it is assumed to be
observable, and one in which it is assumed to be
never observable. This idea is used to refine ICs used to encode the workflow constructs in the case of partial observability.   
For instance, if a partially observable activity appears in the antecedent of an IC, two distinct ICs are generated, one where the activity is considered to be observable ($\happened$), and another in which it is not ($\hyp$).
Thus in the case of a sequence flow from $\act{a}$ to
$\act{b}$, where $\act{a}$ is possibly observable and $\act{b}$ is
observable, IC~\eqref{eq:encoding-sequence} generates:
\[
\begin{array}{r@{~}c@{~}l}
\happened(\act{a},T_a)  &\then& \expf(\act{b},T_b) \land T_b > T_a.\\
\hyp(\act{a},T_a)  &\then& \expf(\act{b},T_b) \land T_b > T_a.
\end{array}
\]
If multiple partially observable activities would appear in the antecedent of an IC (as, e.g., in the and-join case), then all combinations have to be considered.

Similarly, if a partially observable activity appears in the consequent of an IC, a disjunction must be inserted in the consequent, accounting for the two possibilities of observable/never observable event. If both the antecedent and consequent of an IC would contain a partially observable activity, a combination of the rules above will be used.
For example, in the case of a sequence flow from
$\act{a}$ to $\act{b}$, where $\act{b}$ is possibly observable, IC~\eqref{eq:encoding-sequence} generates:
\[
\happened(\act{a},T_a)  \then  
\ought(\act{b},T_b) \land T_b > T_a  
\lor 
\hyp(\act{b},T_b) \land T_b > T_a.
\]
With this encoding, the \sciff\ proof procedure generates firstly an abductive explanation
$\Delta$ containing an expectation about the execution of $\act{b}$. If no
$\act{b}$ is actually observed, $\Delta$ is discarded, and a new
abductive explanation $\Delta'$ is generated
containing the hypothesis about $\act{b}$ (i.e.,
$\hyp(\act{b},T_b)\in\Delta'$). Mutual exclusion between these two
possibilities is guaranteed by the SCIFF declarative semantics (cf.~Definition~\ref{def:fulfillment}).

Finally, if both the antecedent and consequent of an IC would contain a possibly observable activity, a combination of the rules above will be used.

\subsection{Compliance in \sciff: Declarative Semantics}
\label{subsec:sciffSemantics}
We are now ready to provide a formal notion of compliance in its different forms. We do so by extending the \sciff declarative semantics provided in 
\cite{DBLP:journals/tocl/AlbertiCGLMT08} to incorporate log incompleteness (that is, observability features).

A structured process model corresponds to a \sciff specification
$\spec=\langle\mathcal{KB}, \abdset, \ic \rangle$, where:
\begin{inparaenum}[\it (i)]
\item $\mathcal{KB}$ is a Logic Program~\cite{DBLP:books/sp/Lloyd87} containing the definition of accessory predicates;
\item $\abdset = \{\hyp/2,\ought/2\}$;
\item $\ic$ is a set of ICs constructed by following the  encoding defined in Section~\ref{subsec:yawlInSciff}.
\end{inparaenum} A (execution) trace and an abductive explanation $\Delta$ are defined as follows:\footnote{We do not consider the abductive goal, as it is not
  needed for our treatment.}:
  
\begin{definition}
A \emph{Trace} \trace\ is a set of terms of type $\happened(e, T_i)$, where $e$ is a term describing the happened event, and $T_i \in \mathbb{N}$ is the time instant at which the event occurred.
\end{definition}

\begin{definition}[Abductive explanation $\Delta$]
\label{def:explanation}
Given a \sciff\ specification \spec\ and a trace \trace,
a set $\Delta\subseteq\abdset$ is an \emph{abductive explanation} for
$\langle \spec, \trace\rangle$ if and only if
\[
\emph{Comp} \left( \mathcal{KB} \cup \trace \cup \Delta \right) \cup \operatorname{CET} \cup T_\mathbb{N} \models \ic
\]
where \emph{Comp} is the (two-valued) completion of a theory
\cite{DBLP:journals/jlp/Kunen87}, \emph{CET} stands for Clark
Equational Theory \cite{Clark1978:NAF} and $T_\mathbb{N}$ is the CLP
constraint theory \cite{SemanticsCLP} for integers.
\end{definition}



\noindent The following definition fixes the semantics for
observable events, and provides the basis for understanding
the alignment of a trace with a process model.


%
%
\begin{definition}[\trace-Fulfillment]
\label{def:fulfillment}
Given a trace \trace, an abducible set $\Delta$  is
\emph{\trace-fulfilled} if for every event $\act{e} \in \Delta$ and for each
time $\act{t}$, $\ought(\act{e},\act{t}) \in \Delta$ if
and only if $\happened(\act{e},\act{t}) \in \trace$.
\end{definition}
The ``only if'' direction defines the semantics of expectation,
indicating that an expectation is fulfilled when it finds the corresponding happening event 
in the trace. The ``if'' direction captures the prescriptive nature of process
models, whose \emph{closed} nature require that only expected event
may happen.


%
Given an abductive explanation $\Delta$, fulfilment acts as a
\emph{compliance classifier}, which separates the legal/correct execution traces with respect to $\Delta$ from the wrong ones.
\begin{definition}[(Strong/Conditional) Compliance]
\label{def:compliance}
A trace \trace\ is \emph{compliant} with a \sciff\ specification \spec\ if there exists an abducible set $\Delta$ such that:
\begin{inparaenum}[\it (i)]
	\item $\Delta$ is an abductive explanation for $\langle \spec,
          \trace \rangle$, and
	\item $\Delta$ is \trace-fulfilled.
\end{inparaenum}
If $\Delta$ does not contains any \hyp\ then we say that it is \emph{strongly-compliant}, otherwise it is \emph{conditionally-compliant}.
\end{definition}
If no abductive explanation that is also $\trace$-fulfilled can be
found, then $\trace$ is not compliant with the specification of
interest. Contrariwise, the abductive explanation witnesses
compliance. However, it may contain $\hyp$ predicates, abduced due to
the incompleteness of $\trace$. In fact, the presence or absence of
such predicates determines whether $\trace$ is conditionally or
strongly compliant.
To make an example of how Definition~\ref{def:compliance} help us solve the problem of compliance of a single trace, let us consider traces~\eqref{eq:abductedtrace}, \eqref{eq:sample-trace} and \eqref{eq:noncompliant}, of the POS example. In the case of partial trace~\eqref{eq:abductedtrace}, \sciff will tell us that it is conditional compliant with the workflow model POS since $\Delta$ will contain the formal encoding of the two abducibles \eqref{eq:h1} and \eqref{eq:h2} which provide the \emph{abductive explanation} of trace \eqref{eq:abductedtrace}.
In the case of (complete) trace \eqref{eq:sample-trace}, abduction will tell us that it directly follows from $\Gamma$ without the need of any hypothesis. The case $\Delta = \emptyset$ coincides in fact, with the classical notion of (deductive) compliance. Finally, if we consider the partial trace \eqref{eq:noncompliant}
\sciff will tell us that it is not possible to find any set of
hypothesis  $\Delta$ that explains it. The case of no $\Delta$
coincides, therefore with the classical notion of (deductive)
non-compliance.

We close this section by briefly arguing that our approach is indeed
correct. To show correctness, one may proceed in two steps:
\begin{inparaenum}[\it (i)]
\item prove the semantic correctness of the encoding w.r.t.~semantics of
(conditional/strong) compliance;
\item prove the correctness of the proof procedure w.r.t.~the \sciff declarative semantics.
\end{inparaenum}
Step (i) requires to prove that a trace is
(conditionally/strong) compliant (in the original execution semantics of the
workflow) with a given workflow if and only if the trace is (conditionally/strong)
compliant (according to the \sciff
declarative semantics) with the encoding of the workflow in \sciff. This can be done in
the spirit of \cite{Montali2010} (where correctness is proven for
declarative, constraint-based processes), by arguing that structured
processes can be seen as declarative processes that only employ the
``chain-response constraint'' \cite{Montali2010}.
For step (ii), we rely on~\cite{DBLP:journals/tocl/AlbertiCGLMT08}, where t soundness and completeness of \sciff w.r.t. its declarative semantics is proved by
addressing the case of closed workflow models (the trace is closed and no more events can happen anymore), as well as that of open workflow models 
(future events can still happen). Our declarative semantics
restricts the notions of fulfilment and compliance to a specific
current time $t_c$, i.e., to open traces: hence soundness and
completeness still hold.

\subsection{Dealing with Process Execution (In)Completeness in \sciff}
\label{subsec:trace_incompleteness}
We have already illustrated, by means of the POS example, how Definition~\ref{def:compliance} can be used, at a very abstract level, to address compliance of a partial trace. In this section we illustrate more in detail how \sciff can be used to solve the three domensions of incompleteness identified in Section~\ref{sec:classifying_log_incompleteness}.

Trace and event incompleteness are dealt by with \sciff in a uniform manner. In fact, the \textbf{trace/event incompleteness problem} amounts to check if a given log (possibly equipped with incomplete traces/events), is compliant with a prescriptive process model. We consider as input the process model, together with information about the observability of its activities, a trace, and a maximum length for completed traces. The compliance is determined by executing the \sciff\ proof procedure and evaluating possible abductive answers. We proceed as follows:
\begin{enumerate}
\item We automatically translate the process model with its observability
  meta-information into a \sciff\ specification.
If observability information is missing
  for some/all the activities, we can safely assume that some/all
  activities are possibly observable.
\item The \sciff\ proof procedure is applied to the \sciff\ specification
 and to the trace under observation,
computing \emph{all} the possible abductive answers $\Delta_i$. The
 maximum trace length information is used to limit the search, as in
 the unrestricted case the presence of loop may lead to nontermination.
%
\item If no abductive answer is generated, the trace is deemed as
  non-compliant.
 Otherwise, a set of abductive answers $\{\Delta_1,\ldots,\Delta_n\}$
  has been found. If there exists a $\Delta_i$ that does not
  contain any $\hyp$ predicate, then the trace is strongly
  compliant. The trace is conditionally compliant otherwise.
\end{enumerate}
Note that, assessing strong/conditional compliance requires the computation of 
\emph{all} the abductive answers, thus affecting the performances of the \sciff proof procedure. 
If only compliance is needed (without classifying it in strong or conditional), it is possible to compute only the first solution. 

A different scenario is provided by the \textbf{log incompleteness problem}, which instead focuses on an entire event log, and looks if some possible traces allowed by the model are indeed
missing in the log. In this case we consider as input the process
model, a maximum length for the completed traces, and a log consisting of a
number of different traces; we assume each trace is trace- and
event-complete. We proceed as follows:
\begin{enumerate}
\item We generate the \sciff\ specification from the process model,
  considering all activities as never observable.
%
\item The \sciff\ proof procedure is applied to the \sciff\ specification.
\emph{All} the possible abductive answers $\Delta_i$ are computed, with
 maximum trace length as specified. Each answer corresponds to a
 different execution instance allowed by the model.
Since all the activities are never observable, the generated
$\Delta_i$ will contain only \hyp.
\item For each hypothesised trace in the set
  $\{\Delta_1,\ldots,\Delta_n\}$, a corresponding, distinct trace is
  looked for in the log. If all the hypothesised traces have a
  distinct matching observed trace, then the log is deemed as
  complete.
\end{enumerate}
Notice that, beside the completeness of the log, the proof procedure
also generate the missing traces, defined as the $\Delta_i$ that do not have a corresponding trace in the log.

\section{Evaluation}
\label{sec:evaluation}



\begin{figure}[t]
\centering
\includegraphics[width=1.0\linewidth]{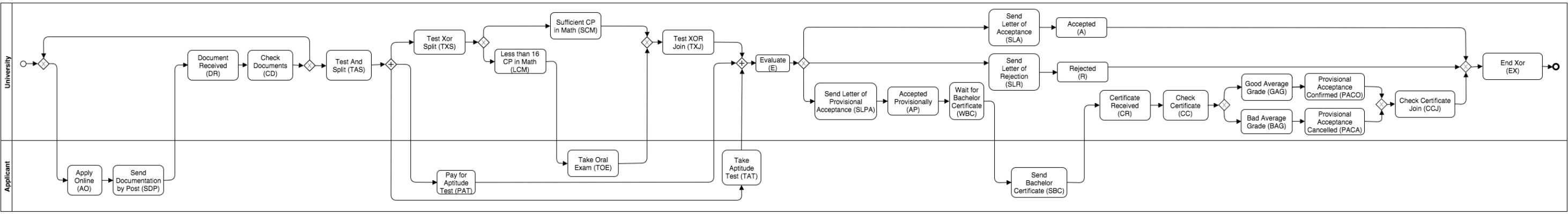}
\vspace{-.5cm}
\caption{University admission process}
\vspace{-.5cm}
\label{fig:university_exam}
\end{figure}

Section~\ref{subsec:trace_incompleteness} illustrates how the eight problems obtained by combining the three incompleteness dimensions can be actually solved by means of two algorithms. We now test such algorithms and study how different inputs affect their performances.
%
As special input, we indicate whether \sciff must compute all possible abductive explanations, or a simple yes/no answer to the compliance decision problem suffices (the latter can be answered affirmatively by stopping after having found the first abductive explanation).

For each type of incompleteness we consider, possibly only some input parameters are of interest, as, for instance, the information on the observability of activities does not impact the log incompleteness resolution (being each trace in the log assumed complete). Hence, for each problem we select the significant parameters only and perform tests by varying them in order to thoroughly understand their practical influence.

As for the model, we choose a real-life process made available within the Process Matching Contest 2013~\cite{PMContest2013}, describing the admission procedure to the Frankfurt University. Notably, in order to exercise the encoding on various process elements, a parallel branch and a loop have been added to the original procedure. Figure~\ref{fig:university_exam} shows the resulting model\footnote{The model is included for providing an intuition of its complexity and no description is provided.}, which is composed of 29 activities, 3 xor-splits/joins, and 1 and-split/join. If no loop iteration is considered, the model contains 8 distinct paths.
The experiments have been carried out on a Windows 7 pc with 8GB RAM and a 2.4 GhZ Intel-core i7.

\setlength{\aboverulesep}{0pt}
\setlength{\belowrulesep}{0pt}
\setlength{\extrarowheight}{.75ex}
\definecolor{Gray}{gray}{0.9}
\newcolumntype{a}{>{\columncolor{Gray}}c}
\begin{table}[t]
  \centering
    \begin{scriptsize}
  \begin{tabular}{|a>{\hspace{0.25em}} a>{\hspace{1em}} a|>{\hspace{0.25em}} c>{\hspace{0.25em}} c>{\hspace{0.25em}} c>{\hspace{0.25em}} l| c>{\hspace{0.25em}} c>{\hspace{0.25em}} c>{\hspace{0.25em}} l|}
  \toprule
     \multicolumn{3}{|a}{\textbf{Input}}  & \multicolumn{4}{c}{\textbf{Output Trace Incompl.}} & \multicolumn{4}{c|}{\textbf{Output Trace and Event Incompl.}}\\
     \textbf{\% AOA} & $\#$ \textbf{OE} & \textbf{TML} & \textbf{Cc} & \textbf{CcT (ms) } & $\#$ \textbf{sol.}  & \textbf{CT (ms)} & \textbf{Cc} & \textbf{CcT (ms) } & $\#$ \textbf{sol.}  & \textbf{CT (ms)} \\ \midrule
  & & 16 & YES & 78 & 2 & 213 & YES  & 234 & 28 & 7069 \\
    & \multirow{-2}{*}{1}& 32 & YES & 109 & 16 & 915 & YES & 219  & 694 & 36754 \\ \cmidrule(l){2-11}
    & & 16 & NO & 551 & 0 & 551 & NO & 24209 & 0 & 24209 \\ 
    & \multirow{-2}{*}{5} & 32 & YES &  $>4h$ & 6 & 6759 & YES & $>4h$ & 134 &270278 \\ \cmidrule(l){2-11}
    & & 16 & NO & 38966 & 0 & 38966 & NO & 2139880 & 0 & 2139880\\ 
    \multirow{-6}{*}{0} & \multirow{-2}{*}{15} & 32 & YES & 683073 & 36 & 13599417 &  &  $>4h$ &  & $>4h$ \\ \midrule
    & & 16 & NO & 266 & 0 & 20 & YES & 266 & 2 & 124 \\
    & \multirow{-2}{*}{1}& 32 & NO & 16 & 0 & 16 & YES & 234 & 2 & 405\\ \cmidrule(l){2-11}
    & & 16 & NO & 548 & 0 & 548 & YES & 3151 & 1 & 3650 \\
     & \multirow{-2}{*}{5} & 32 & YES & 124 & 1 & 157 & YES & 2964 & 1 & 4493 \\ \cmidrule(l){2-11}
    & & 16 & NO & 43209 & 0 & 43209 & YES & 255483 & 1 & 28980 \\
     \multirow{-6}{*}{15\%}& \multirow{-2}{*}{15}  & 32 & YES & 25210 & 36 & 13257351 & YES & 255483 & 1 & 245810 \\ \midrule
    & & 16 & NO & 31 & 0 & 31 & NO & 156 & 0 & 156\\
    & \multirow{-2}{*}{1}& 32 & NO & 31 & 0 & 31 & NO & 187 & 0 & 187 \\ \cmidrule(l){2-11}
    & & 16 & NO & 16 & 0 & 16 & NO & 202 & 0 & 202 \\
     & \multirow{-2}{*}{5} & 30 & NO & 31 & 0 & 31 & NO & 256 & 0 & 256\\ \cmidrule(l){2-11}
    & & 16 & YES & 874 & 1 & 999 & YES & 9734 & 1 & 8346\\
    \multirow{-6}{*}{$~$ 50\%}& \multirow{-2}{*}{15} & 32 & YES & 827 & 1 & 1123 & YES & 12730 & 1 & 9812 \\ 
    \bottomrule
  \end{tabular}
  \caption{Results related to the trace and trace and event incompleteness.}
  \label{tab:tr_ev_results}
  \end{scriptsize}
  \vspace{-.8cm}
\end{table}

\noindent
\textbf{Log incompleteness.}
We evaluated the algorithm by varying the number of (complete) traces in the log and the bound on the length of the traces. Results shows that the number of solutions is proportional to the trace max length parameter and inversely proportional to the number of traces already presented in the log. Computing times are below 1 sec, and proportional to the number of solutions returned.

\noindent
\textbf{Trace/event incompleteness.}
We first feed the algorithm with a partial trace and complete events, thus testing the trace incompleteness problem. 
Table~\ref{tab:tr_ev_results} is used to summarize the numerical values obtained, where each row represents a test case and columns report the input values used, the output and the computing time. As for the activity observability, we chose as parameter the percentage of activities that we know to be always/never observable ($\% AOA$) and let it ranging among $0\%$ (no certain information about activity observability is available at all), around $15\%$ ( observability is known with certainty only for a small number of activities) and $50\%$ (for about half of the activities of the diagram we know whether they are for sure observable or not). Concerning the trace, we look at the number of observed events in the trace ($\# OE$).
Also in this case, we chose to make the parameter varying among a trace almost incomplete (1), with a small number of observed events (5) and with a medium number of observed events (15). Finally, we let the trace maximum length (TML) ranging among the values 16 (shortest path without loop) and 32 (up to 2 loops).

The table shows that the computation time when all possible completions are returned (CT) is proportional to the bound on the length of the traces (TML) and the number of completions found (\#sol). Indeed, when one or more solutions are found, the computation time for the compliance decision problem ($CcT$) is roughly $CT / \#sol$.
On the contrary, the percentage of always observed activities in the model ($\%AOA$) significantly reduces the exploration space, thus linearly decreasing the computation time.
Interestingly, the more events we observe in the trace (column $\# OE$) the higher the computation time is. This is because, as explained in Subsection~\ref{subsec:trace_incompleteness}, for each expectation about a (possibly observable) activity in the model which can potentially match (unify with) an event observed in the trace, two cases must be explored: either (i) the expectation is matched with the event in the trace, or (ii) an abductive explanation is generated for that activity. We remark that the presence of loops in the model requires both the alternatives to be explored, if we want to guarantee that \emph{all} completions are returned.

The rightmost part of Table~\ref{tab:tr_ev_results} shows the results obtained by feeding the procedure with an incomplete trace containing an incomplete event. Here, the same considerations of the trace incompleteness test described above can be drawn. The only difference lies in the fact that a high number of incomplete events (i.e., events lacking their description and, in particular, their name) let the computation time rise exponentially, as multiple possibilities for each incomplete event must be explored when looking for the set of all possible completions. Again, this is the price we pay to get the completeness of the results, which however can (in general) be avoided by asking for compliance only, as displayed by column $CnT$ in Table~\ref{tab:tr_ev_results}.

We close the section by analyzing event incompleteness, that is a particular case of the trace and event incompleteness described above. In this case, the incompleteness is on the event description only (see Section~\ref{sec:classifying_log_incompleteness}), hence an input trace is characterized by a number of observed events equals to the length of the trace, and, among these, a number of missing event descriptions. From the computational viewpoint, this represents the most challenging setting, as both the above parameters cause an exponential increase in the computation times. We reckoned that for an input trace of length $16$, even two missed event description brings the computation time up to a couple of hours.

\subsection{Discussion}
The purpose of the experimentation was to stress the algorithm in borderline cases. Indeed, we remark that on the one hand we made the model convoluted on purpose especially by adding a loop, which is a source of complexity, and on the other we tested situations that are very unlikely to happen in practice. We can safely assume that in typical scenarios the number of (partially) observable activities, which in most of the cases are human-performed, is usually no more than half of the overall activities (i.e., $\%AOA > 50\%$) and that from the $\%AOA$ parameter and the length of the partial trace, a good estimate of the bound on the trace length can be set, thus avoiding useless loop (when present) iterations. In such settings the performance of the abductive procedure on the different types of incompleteness are reasonable. For instance, for the compliance, they range from few seconds when at most a single event description is completely unknown to about $4.5$ minutes when up to $4$ event descriptions are missing.

\section{Related Work}
\label{sec:related_works}

The problem of incomplete traces has been tackled by a number of works in the field of process discovery and conformance. 
%
Some of them~\cite{Adriansyahetal:2011,deLeonietal:2012} have addressed the problem of aligning event logs and procedural/declarative process models~\cite{Adriansyahetal:2011,deLeonietal:2012}. 
Such works explore the search space of the set of possible moves to find the best one for aligning the log to the model.
Our purpose is not managing generic misalignments between models and logs, but rather focus on a special type of incompleteness: the model is correct and the log could be incomplete. 

%
%
%
We can divide existing works that aim at constructing possible model-compliant ``worlds'' out of a set of observations with incomplete information in two groups: quantitative and qualitative approaches.
The former rely on the availability of a probabilistic model of execution and knowledge. 
For example, in~\cite{Rogge-Solti2013} the authors exploit stochastic Petri nets and Bayesian Networks to recover missing information. 
The latter stand on the idea of describing ``possible outcomes'' regardless of likelihood. 
Among these approaches, the issue of reconstructing missing information has been tackled in~\cite{Bertoli2013} and in~\cite{CAiSE2015}, respectively leveraging Satisfiability Modulo Theory, and planning techniques.

In this work, the notion of incompleteness has been investigated and extended to take into account its different variants (\textit{log incompleteness}, \textit{trace incompleteness} and \textit{event incompleteness}). Similarly, the concept of \emph{observability} has been deeper investigated and extended, by exploring the case of activities \textit{always}, \textit{partially} or \textit{never} observable. This has led to a novel classification of different ``degrees'' of compliance. 


Abduction and the \sciff\ framework have been previously used to model both procedural and declarative processes. In \cite{DBLP:conf/aime/ChesaniMMS07}, a structured workflow language has been defined, with a formal semantics in \sciff. In \cite{DBLP:journals/tweb/MontaliPACMS10}, \sciff has been exploited to formalize and reason about the declarative workflow language Declare.

An interesting work where trace compliance is evaluated through abduction is presented in \cite{Zannone2015}. However, they define compliance as assessing if actions were executed by users with the right permissions (auditing), and focus only on incomplete traces (with complete events), while we take a more sophisticated approach to incompleteness. The adopted abductive framework, CIFF \cite{DBLP:journals/tplp/MancarellaTSTE09}, only supports ground abducibles, and ICs are limited to denials. The work in \cite{Zannone2015} explores also the dimension of human confirmation of hypotheses, and proposes a human-based refinement cycle. This is a complementary step with our work, and would be an interesting direction for future work.


\section{Conclusions}
\label{sec:conclusion}

We have presented an abductive  framework to support business process monitoring (and in particular compliance checking) by attacking the different forms of incompleteness that may be present in an event log. 
%
Concerning future development, the \sciff framework is based on first-order logic, thus paving the way towards (i) the incorporation of data~\cite{DBLP:conf/ispw/MasellisMM14}, (ii) extensions to further types of workflows (e.g., temporal workflows as in~\cite{KumarBPM2015}), and (iii) towards the investigation of probabilistic models to deal with incompleteness of knowledge. 

\bibliographystyle{splncs}
\bibliography{BPM2016-bib}

\end{document}